\documentclass[10pt, conference]{IEEEtran}
\IEEEoverridecommandlockouts
\usepackage{cite}
\usepackage{amsmath,amssymb,amsfonts}
\usepackage{enumitem}
\usepackage{algorithm}
\usepackage{algpseudocode}
\usepackage{graphicx}
\usepackage{hyperref}
\usepackage{textcomp}
\usepackage{xcolor}
\algnewcommand{\Compute}{\textbf{Compute:}}
\algnewcommand{\Input}{\textbf{Input:}}
\algnewcommand{\Output}{\textbf{Output:}}

\def\BibTeX{{\rm B\kern-.05em{\sc i\kern-.025em b}\kern-.08em
    T\kern-.1667em\lower.7ex\hbox{E}\kern-.125emX}}
\begin{document}

\title{Explaining Deep Neural Networks for Bearing Fault Detection with Vibration Concepts}

\author{\IEEEauthorblockN{Thomas Decker\textsuperscript{\rm 1,}\textsuperscript{\rm 2}, Michael Lebacher\textsuperscript{\rm 2}, Volker Tresp\textsuperscript{\rm 1,}\textsuperscript{\rm 2}\\
\textsuperscript{\rm 1}Ludwig Maximilians Universität,
Munich, Germany and \textsuperscript{\rm 2}Siemens AG, Munich, Germany\\ \{thomas.decker, michael.lebacher, volker.tresp\}@siemens.com}}

\maketitle

\begin{abstract}
Concept-based explanation methods, such as Concept Activation Vectors, are potent means to quantify how abstract or high-level characteristics of input data influence the predictions of complex deep neural networks. However, applying them to industrial prediction problems is challenging as it is not immediately clear how to define and access appropriate concepts for individual use cases and specific data types. In this work, we investigate how to leverage established concept-based explanation techniques in the context of bearing fault detection with deep neural networks trained on vibration signals. Since bearings are prevalent in almost every rotating equipment, ensuring the reliability of intransparent fault detection models is crucial to prevent costly repairs and downtimes of industrial machinery. Our evaluations demonstrate that explaining opaque models in terms of vibration concepts enables human-comprehensible and intuitive insights about their inner workings, but the underlying assumptions need to be carefully validated first.
\end{abstract}


\section{Introduction}
\begin{figure}[t]
	\centering
	\includegraphics[width=0.99\columnwidth]{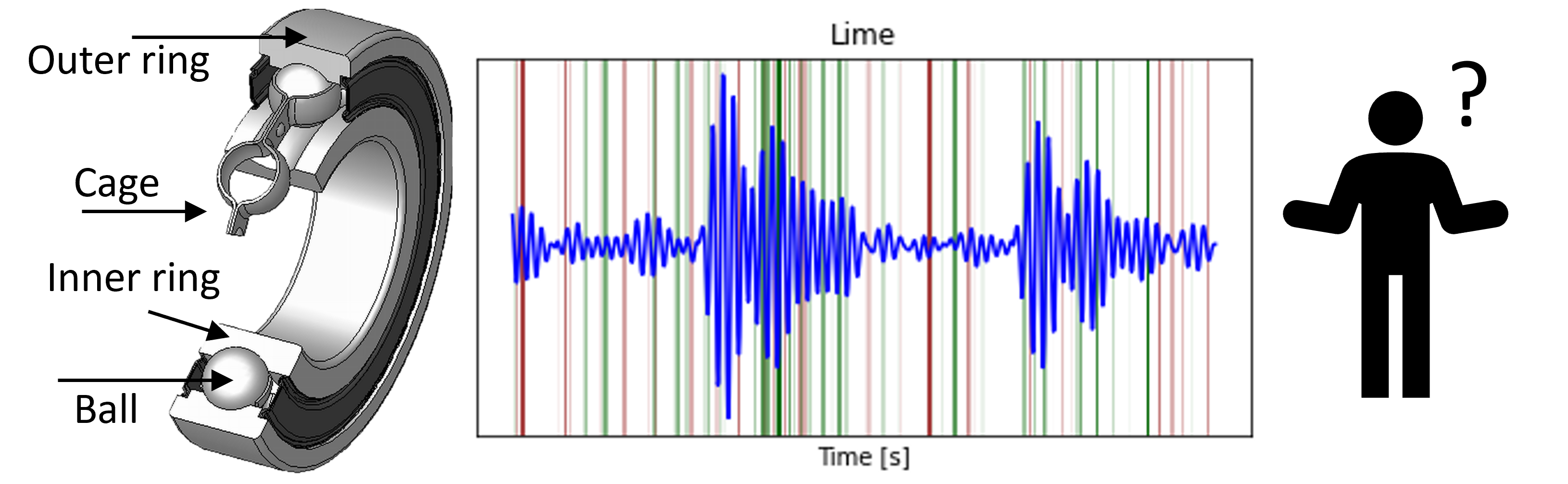}
	\caption{Left: Schematics of a rolling element bearing and its components. Image Credit: Silberwolf / CC-BY-2.5. Right: Incomprehensible results of LIME highlighting the importance of signal values for the prediction of a deep neural network. Explanations in this form are hard to interpret for humans. }
	\label{fig:1}
\end{figure}

Bearings (Fig \ref{fig:1}) are critical mechanical components that are omnipresent in many industrial applications. They are used to support and reduce friction in rotating shafts, which are commonly found in motors, pumps, fans, and other types of rotating equipment. But over time, normal wear and tear or exogenous factors like improper lubrication or excessive overloading can cause bearings to degrade. Unexposed faults of this kind will not only lead to higher maintenance efforts and losses in productivity but might additionally compromise safe machine operation in general. To prevent such scenarios, various methods have been proposed to enable fast and reliable bearing fault detection using artificial intelligence \cite{liu2018artificial}. Such methods are typically trained to identify and analyze patterns in vibration signals obtained during machine operation. Implementing fault detection using deep neural networks has demonstrated very promising capabilities \cite{jia2016deep} as such models are able to achieve impressive performance on various benchmark datasets \cite{zhao2020deep}. Nevertheless, deep neural networks are also considered black box models as their complicated computational structure makes it infeasible to fully comprehend their prediction logic. This makes it difficult to rigorously assess their trustworthiness and reliability when facing real-world circumstances under actual deployment. To overcome these inherent limitations, various techniques have been developed to make such models more transparent to humans, which are also referred to as Explainable AI \cite{arrieta2020explainable}. Some prominent methods include for instance LIME \cite{ribeiro2016should}, SHAP \cite{lundberg2017unified}, or gradient-based approach \cite{ancona2018towards}, which all aim to explain single model prediction by evaluating the influence of individual input features. However, such methods usually indicate feature importance in raw units of the model input domain. For deep neural networks trained on raw vibration signals, this would merely indicate how individual signal values in time affect the model output, which is also illustrated in Fig \ref{fig:1}. Formulating a concrete reason that reveals why the model has detected an outer ring fault based on signal value importance is challenging as such kinds of explanations do not convey insights with an appropriate degree of abstraction and generality. While there exist techniques to adapt such methods to this particular use case \cite{decker2023does} another appealing remedy is provided by concept-based explanations \cite{yeh2022human}.
This category of explanation methods enables quantifying how important high-level and potentially abstract characteristics of input data are for model predictions. For instance, Concept Activation Vectors (CAVs)\cite{kim2018interpretability} have been used to analyze if the presence of the concept "stripes" influences an image classifier when detecting zebras. However, defining appropriate concepts and conducting corresponding analyses for industrial vibration signals is not straightforward and requires adequate design choices grounded in problem-specific domain expertise. In this work, we investigate the applicability and utility of concept-based explanations in the context of bearing fault detection. This is to the best of our knowledge the first attempt to rigorously apply such techniques in the context of vibration signal analysis related to industrial applications. Our contributions are threefold: 
\begin{itemize}
    \item We introduce simulated high-frequency resonances as a meaningful way to exemplify important high-level signal characteristics related to bearing fault vibrations.
    \item We utilize these vibration concepts to derive corresponding Concept Activation Vectors for the purpose of model explanation for different fault detection architectures.
    \item We analyze the faithfulness as well as the utility of concept-based explanations in the context of bearing fault detection and demonstrate how to derive novel insights. 
\end{itemize}
\section{Background}
\subsection{Bearing Fault induced Vibrations}
Defective bearings emit characteristic vibrations during machine operation depending on the precise location of the fault, which could, for instance, occur at the inner or outer ring. This can be illustrated by modeling the bearing and connected parts as a simple linear system that repeatedly gets excited during rotation \cite{randall2011rolling}. Whenever a ball hits the defect located at one of the rings, an impact gets induced that causes vibrations in the form of high-frequency resonances of connected machine parts. This can be expressed via a periodic impulse train denoted by $i(t) = \sum_{k=-\infty}^{\infty} a(t)\delta(t-kT)$, where $\delta$ describes the discrete unit impulse function, $T$ the time between impulses and $a(t)$ the amplitude of each impulse, which depends on parameters like fault size or load distribution. Note, that the frequency of impacts is characteristic of the fault origin and can be computed explicitly based on the bearing's geometry \cite{randall2011rolling}. For outer and inner ring faults these frequencies are referred to as the Ball Passing Frequency Outer ring (BPFO) and Ball Passing Frequency Inner ring (BPFI). Given the bearing diameter $D$, ball diameter $d$, number of balls $n$, and the bearing contact angle $\alpha$, such frequencies can be expressed in dependence of the bearing's rotation frequency $f_r$: 
\begin{align*}
    \textit{BPFO}(f_r) &= \frac{n}{2}f_r\left(1- \frac{d}{D} \cos \alpha \right)\\
    \textit{BPFI}(f_r) &= \frac{n}{2}f_r\left(1+ \frac{d}{D}\cos \alpha \right)
\end{align*}
Thus, the occurrence of a fault at a specific position inside the bearing will lead to periodic impact impulses with an indicative frequency. Moreover, each impulse will excite the system causing an impulse response denoted by $h(t)$. This response $h(t)$ will typically consist of high-frequency resonances of the structure with exponentially decaying magnitude in time \cite{mcfadden1984model}. The resulting fault-induced vibrations can simply be modeled as system output caused by $i(t)$ resulting from convolution with the corresponding impulse response function: $x_{\textit{fault}}(t) = h(t)\ast i(t)$, which can be recorded by a vibration sensor. Although rather simple, this signal model is able to express important physical properties of bearing fault-induced vibrations that are helpful to detect and distinguish different fault types in practice \cite{randall2011rolling}. A comprehensive discussion and other approaches to model such signals can be found in \cite{singh2015extensive}.  
\begin{figure*}[!t]
	\centering
	\includegraphics[width=0.90\textwidth]{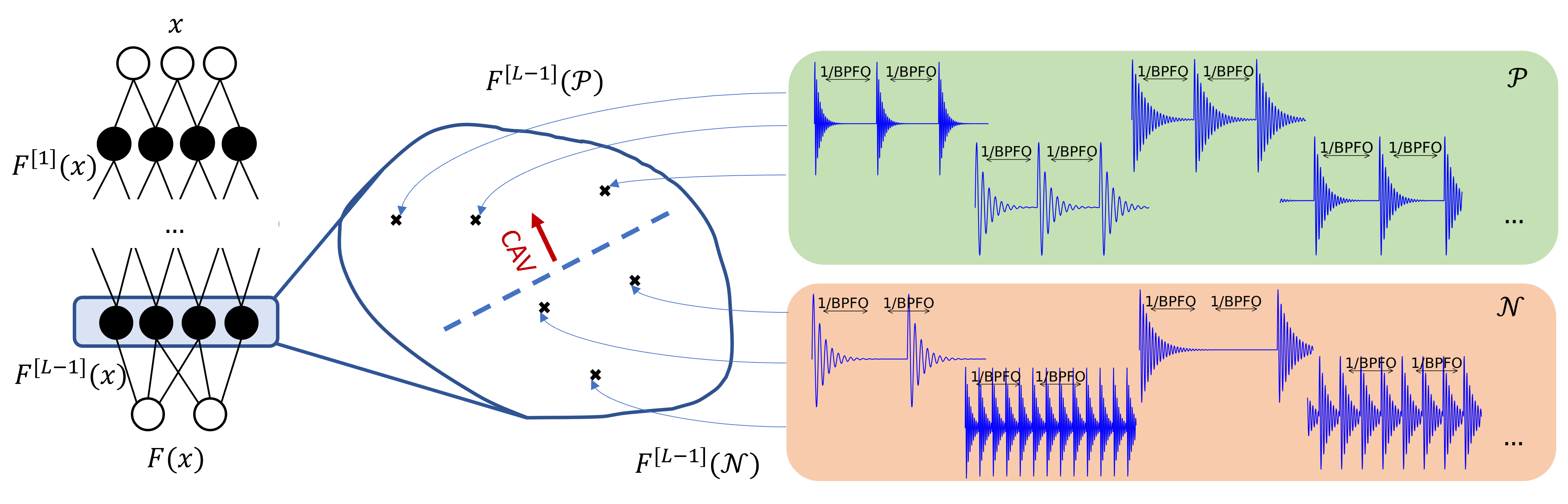}
	\caption{Obtaining Concept Activation Vectors (CAVs) \cite{kim2018interpretability} for vibration concepts to explain a deep neural network $F$ trained to perform bearing fault detection on raw vibration signals $x$: Our approach allows us to define and simulate meaningful vibration concepts that express relevant but high-level signal properties grounded in domain knowledge. It starts by simulating a set of diverse positive examples $\mathcal{P}$  where all impacts occur with the characteristic frequency $\textit{BPFO}$ and a set of negative ones $\mathcal{N}$ where the occurrence is random. These sets can be used to localize the concepts of ''being modulated with frequency $\textit{BPFO}$'' in the network's activation space of layer $L-1$ by training a linear classifier to separate $F^{[L-1]}(\mathcal{P})$ from $F^{[L-1]}(\mathcal{N})$. The direction oriented towards $F^{[L-1]}(\mathcal{P})$ defines the corresponding CAV which can be used to evaluate how important the concept is for the predictions of the deep neural network $F$. }
	\label{fig:2}
\end{figure*}
\subsection{Concept-based Explanations with TCAV}
The goal of concept-based explanations is to enable insights into how a machine learning model forms its predictions in terms of comprehensible and intuitive concepts. In this context, a concept can be understood as any human understandable characteristic related to the data domain of interest. When dealing with image data, such concepts could, for instance, correspond to the visual appearance of different textures, shapes, or colors. Explaining an image classifier based on such concepts could for example answer concrete questions such as "does the presence of stripes matter for an image to be classified as a zebra", or "is the color red important for a model to detect fire engines?" 
\cite{kim2018interpretability}. Testing with Concept Activation Vectors (TCAV) \cite{kim2018interpretability} is a popular method for neural networks that globally quantifies the sensitivity of a classwise prediction with respect to a predefined concept. It assumes that different concepts are linearly separable in the network's activation space and requires access to suitable example data which resemble the concepts of interest. To mathematically derive TCAV scores consider the following setup. Let $F:\mathbb{R}^d \rightarrow \mathbb{R}^C$ be a deep neural network trained to map $d$-dimensional inputs $x$ to $C$-dimensional output scores such that $F_c(x) \in \mathbb{R}$ is indicative for predicting class $c$. Deep neural networks are typically structured in terms of computation layers such that $F(x)=F^L \circ \dots \circ F^1(x)$ and $F^{[l]}(x) =F^l \circ \dots \circ F^1(x)$ describes the activation vector of the $l$-th layer given input $x$.
TCAV associates each concept of interest with a corresponding Concept Activation Vector $\textit{CAV}$ that characterizes the concept in terms of the network's activation patterns after a specific layer. To do so it requires access to a set of positive examples inputs $\mathcal{P}\subset \mathbb{R}^d$ which exhibit the concept and a set of negative examples $\mathcal{N}\subset \mathbb{R}^d$ where the concept is absent. Based on these exemplifying datasets, a binary linear classifier can be trained to separate the corresponding activations $F^{[l]}(\mathcal{P})$ and $F^{[l]}(\mathcal{N})$ at a predefined layer $l$. Given such a linear discriminant, the CAV of a concept is defined as the normal vector oriented toward the activation values of positive examples. Thus, CAVs express directions in the network's activation space that capture the common characteristics of the underlying examples exhibiting a high-level concept. Therefore, CAVs can be used to evaluate the sensitivity of a class prediction $F_c(x)$ concerning a concept by evaluating the directional derivative with respect to the activation values along its CAV: $\nabla_{F^{[l]}} F_c(x) \cdot \textit{CAV}$. This estimates how the prediction of a certain class score changes if a concept would be marginally more present in the input $x$. The final TCAV score is computed by aggregating concept sensitivities over a set of input examples $\mathcal{X}\subset \mathbb{R}^d$ where the importance of the concept should be evaluated. One common way to do so is by measuring the share of examples in $\mathcal{X}$ with positive concept sensitivity:
\begin{align*}
    \textit{TCAV}_c^l = \lvert \left\{ x \in \mathcal{X} : \nabla_{F^{[l]}} F_c(x) \cdot \textit{CAV} > 0 \right\} \rvert / \lvert \mathcal{X} \rvert
\end{align*}
As concepts tend to become more separable at later layers \cite{kim2018interpretability}, we consider TCAV scores always with respect to the layer $L-1$ before the final output is computed so $\textit{TCAV}_c = \textit{TCAV}_c^{L-1}$. By repeating TCAV computation over multiple sets of positive and negative examples for the same concept one is also able to derive confidence intervals and conduct tests assessing the statistical significance of the obtained scores \cite{kim2018interpretability}. By now, there exist also many variations of this technique to derive concept-based explanations \cite{yeh2020completeness, schrouff2021best, achtibat2022towards, crabbe2022concept, bai2023concept} but we only rely solely on the original formulation for the remainder of this paper. Concept-based explanations using TCAV have been already successfully applied to diverse applications, like meteorology\cite{sprague2019interpretable}, medical imaging \cite{clough2019global}, emotion recognition \cite{asokan2022interpretability}, recommender systems \cite{gopfert2022discovering}, natural language processing \cite{nejadgholi2022improving} or electronic health records \cite{mincu2021concept}.

\section{Defining suitable vibration concepts}

In order to define adequate concepts for the purpose of model explanations, the authors in \cite{ghorbani2019towards} propose three necessary prerequisites. First, concepts and exemplifying examples shall be intrinsically meaningful to humans on their own. Second, examples exhibiting a concept should all be coherent in sharing common properties that can be associated with the concept while still being clearly distinguishable from other concepts. Third, concepts need to be important and relevant for the prediction problem to be analyzed. Guided by these desiderata, we propose to define 
vibration concepts as simulated high-frequency resonances that are amplitude-modulated with a distinctive frequency. Remember that these kinds of signals express expected properties related to the expected structure of simplified bearing fault vibrations. The computational procedure that we use to simulate such vibration concepts is described in Algorithm \ref{alg:vib} and some noise-free examples of simulated concepts are depicted in Fig. \ref{fig:2}. Note, that each simulated signal is meaningful on its own as it represents an idealized version of a potential bearing fault-related vibration. Examples of such vibration concepts are also coherent as they all share the common feature of periodic modulation with the same characteristic frequency while still being individually different for varying choices of the remaining parameters. Furthermore, they are also important for the task of bearing fault detection as the presence of specifically modulated resonances is a typical indicator of the presence of a certain fault type.  
\begin{algorithm}
\caption{Simulating examples of vibration concepts}\label{alg:vib}
\begin{algorithmic}
\State \textbf{Input:} $\;\;$ characteristic fault frequency $f_{\textit{char}}$,\newline \indent\indent  resonance frequency  $f_{\textit{res}}$,  amplitude $a$, \newline \indent\indent time decay $\tau$, noise level $\sigma$, offset $t_0$ \newline
\State \textbf{Compute:}  $\; i(t)= \sum_{k} a\delta(t-k/f_{\textit{char}}+t_0)$\\
  \indent\indent\indent   $\quad h(t)= e^{-t/\tau} cos(2\pi f_{\textit{res}}t)$\\
  \indent\indent \indent  $\quad \varepsilon(t) \sim \mathcal{N}(0, \sigma^2)$\newline

\State \textbf{Output:} $\quad x_{vib} = h(t)\ast i(t) + \varepsilon(t)$
\end{algorithmic}
\end{algorithm}

\section{Explaining in terms of Vibration Concepts}
To overcome the limitations of common feature importance techniques when applied to fault detection models for raw vibration data, we propose to rather analyze such models in terms of the above-defined vibration concepts. This is also illustrated in Fig. \ref{fig:2}. Consider, for instance, a set of vibration signals $\mathcal{X}_{\textit{outer}}^{f_r}$ originating from an outer ring fault during rotation with frequency $f_r$. To better understand how a deep neural network performs fault classification, a domain expert might explicitly seek to assess whether the model is sensitive to the presence of high-frequency resonances modulated with $\textit{BPFO}(f_r)$ when evaluating signals in $\mathcal{X}_{\textit{outer}}^{f_r}$. This would be a desirable model behavior as such sensitivity implies that the model has indeed learned a valid relationship that is in line with existing domain knowledge. In the case of inner fault signals the same reasoning applies with $\textit{BPFI}(f_r)$. Leveraging TCAV in combination with vibration concepts we are now in a position to retrieve such kind of explanations as follows. Consider again the signals $\mathcal{X}_{\textit{outer}}^{f_r}$ introduced above. Utilizing Algorithm \ref{alg:vib} we can simulate a set of positive examples $\mathcal{P}_{\textit{outer}}^{f_r}$, where we fix $f_{char}=\textit{BPFO}(f_r)$ and vary all remaining parameters. To obtain a set of negative examples $\mathcal{N}$ we simulate vibration concepts where we additionally randomize the modulation frequency $f_{char}$ over a specified interval. This ensures that the resulting $\textit{CAV}$ precisely captures the high-level property of being modulated with the right characteristic frequency as this is the only distinguishable property between both sets of concepts. The computed score of this setup $\textit{TCAV}_{\textit{outer}}^{f_r} \in [0,1]$ measures the proportion of outer ring fault vibrations in $\mathcal{X}_{\textit{outer}}^{f_r}$ where the corresponding prediction shows positive sensitivity with respect to the concept.
Note that such kind of information is immediately accessible and comprehensible to a human expert trying to validate and comprehend the prediction logic of an opaque fault detection model. To demonstrate the applicability and utility of this kind of analysis we conducted dedicated numerical experiments.

\section{Numerical Experiments}
We consider two publicly available datasets containing real vibration recordings obtained under varying machine operating conditions. To compute TCAV scores we use Captum \cite{kokhlikyan2020captum}.
\subsection{Dataset details} 
\subsubsection{CWRU}
The Case Western Reserve University (CWRU) \cite{CWRU} provides real vibration recordings of healthy and damaged bearings with respect to various damage types under four different rotation speeds with varying fault sizes and load conditions. For outer ring fault data, also different positions of the damage on the ring are considered. We restrict our analysis to healthy, inner and outer ring damage signals measured with 48kHz at the drive end sensor. All faults are mechanically enforced using electro-discharge machining. More details about this dataset can be found in \cite{smith2015rolling}. We partitioned all available signals into segments of size $d=3000$ with no overlap and normalized them to the range of -1 to 1 to obtain the dataset CWRU3000. The set CWRU9000 was constructed equivalently containing signal of length $d=9000$.
\subsubsection{Paderborn}
The dataset published by Paderborn University \cite{lessmeier2016condition} contains vibration data collected from a test motor running with six different undamaged bearings as well as various kinds of damaged ones. These include inner and outer ring faults that are either artificially induced using different mechanical procedures or generated via accelerated lifetime tests to mimic more realistic damages from actual wear and tear. During the extensive experiments, also varying operating conditions and damage levels have been considered including two different rotation speeds. Precise information about all conducted experiments is documented in \cite{lessmeier2016condition}. For this data source, we divided all signals into non-overlapping segments of size $d=16000$, normalized them, and constructed two separate datasets. PB artificial is formed by all healthy signals and all fault signals that are artificially enforced while PB realistic contains all healthy signals together with all realistic faults resulting from actual degradation.
\begin{figure}[htb]
	\centering
	\includegraphics[width=0.99\columnwidth]{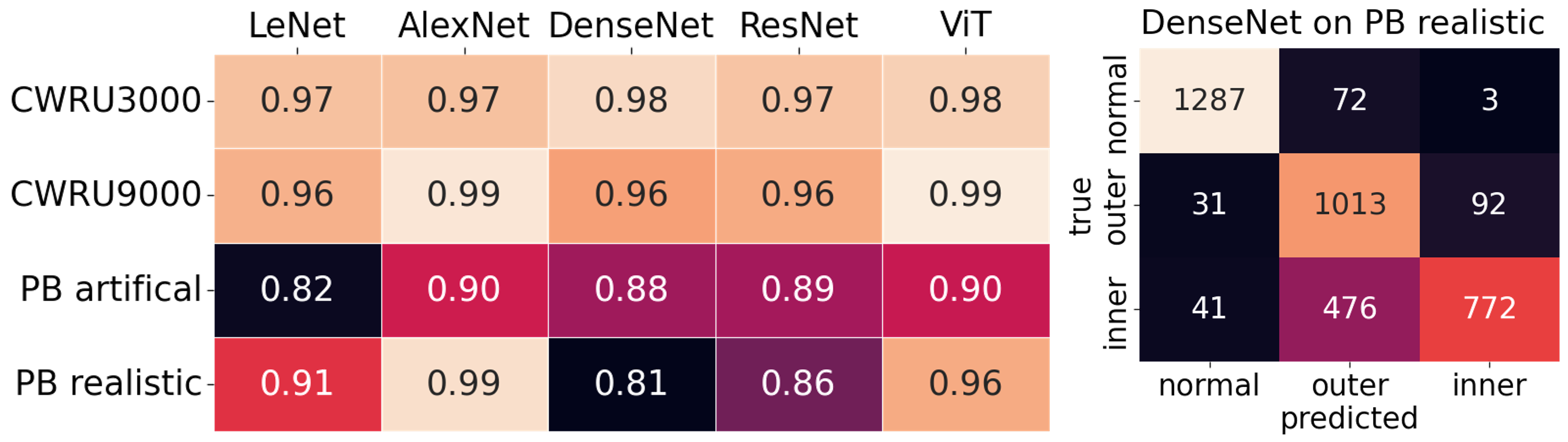}
	\caption{Left: Test accuracies for all considered deep neural network architectures and datasets. Right: Confusion matrix of DenseNet on PB realistic.}
	\label{fig:3}
\end{figure}
\subsection{Fault detection models and training}
To train deep neural networks for bearing fault detection, we consider different architectures that are popular in computer vision. In particular, we fitted one-dimensional versions of LeNet \cite{lecun1998gradient}, AlexNet \cite{krizhevsky2017imagenet}, ResNet \cite{he2016deep} with four residual blocks, DenseNet \cite{huang2017densely} with four densely connected blocks, and Vision Transformers (ViT) \cite{dosovitskiy2020image} with a patch size of 250 to classify vibration signals. Following common practice, we split all datasets into train, validation and test sets with shares 60/20/20. All models are trained for 50 epochs to distinguish between healthy, inner, and outer fault-related signals based on the training set and we selected the final model for each architecture based on the validation loss. The resulting test performances for all final models are presented in Fig. \ref{fig:3} (left). On the CWRU datasets, all networks are able to attain high accuracies independent of the signal length while on Paderborn data the performance tends to be lower, especially for LeNet, DenseNet and ResNet. Moreover, many models show similar prediction capabilities, although our architecture choice reflects various network designs, sizes, and complexities. This raises the question of whether the models also exhibit the same internal reasoning to solve the task. To better understand how the models form their prediction, concept-based explanations can be useful due to their capability to provide relevant insights in human comprehensible form. To assess the applicability and utility of TCAV with vibration concepts for this purpose, we conducted a detailed analysis.

\begin{figure*}[htb]
	\centering
	
 \includegraphics[width=0.54\textwidth]{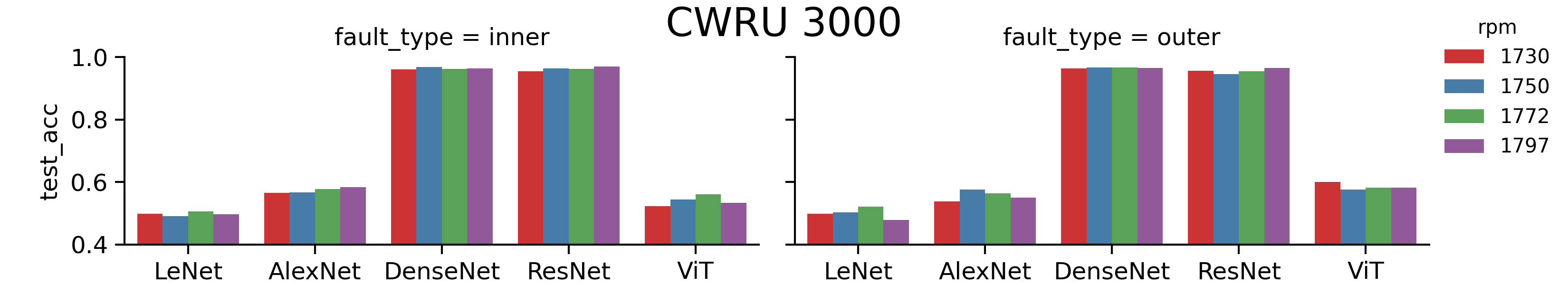}
 \includegraphics[width=0.44\textwidth]{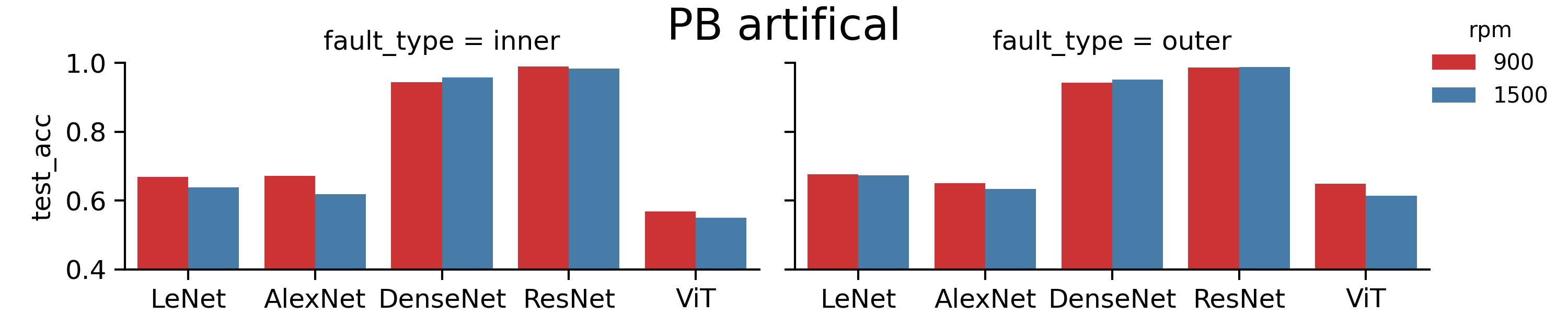}
 \includegraphics[width=0.54\textwidth]{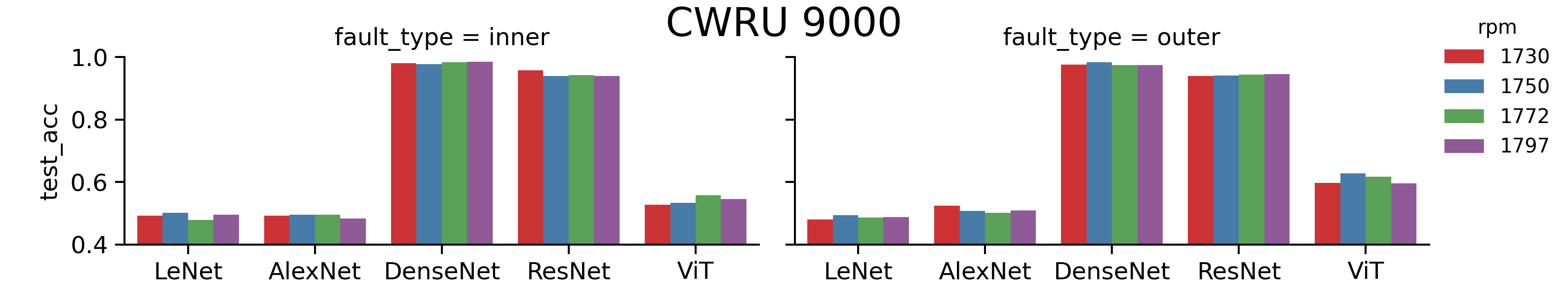}
 \includegraphics[width=0.44\textwidth]{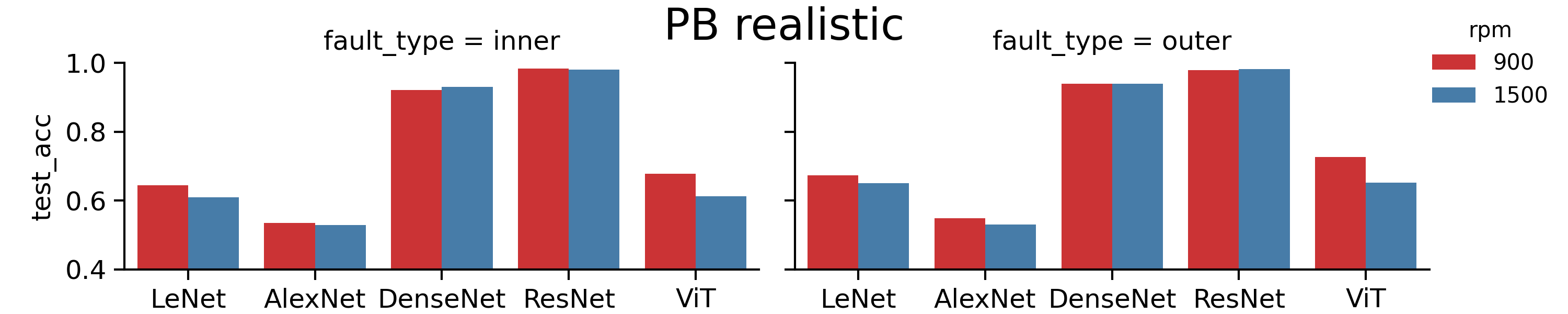}
	\caption{Test accuracies of a linear binary classifier trained to separate vibration concepts in the activation space of different considered model architectures. Only DenseNet and ResNet can achieve consistently good results, indicating that the corresponding TCAV scores are faithful only for such models.}
	\label{fig:4}
\end{figure*}
\begin{figure}[htb]
	\centering
	
  \includegraphics[width=0.80\columnwidth]{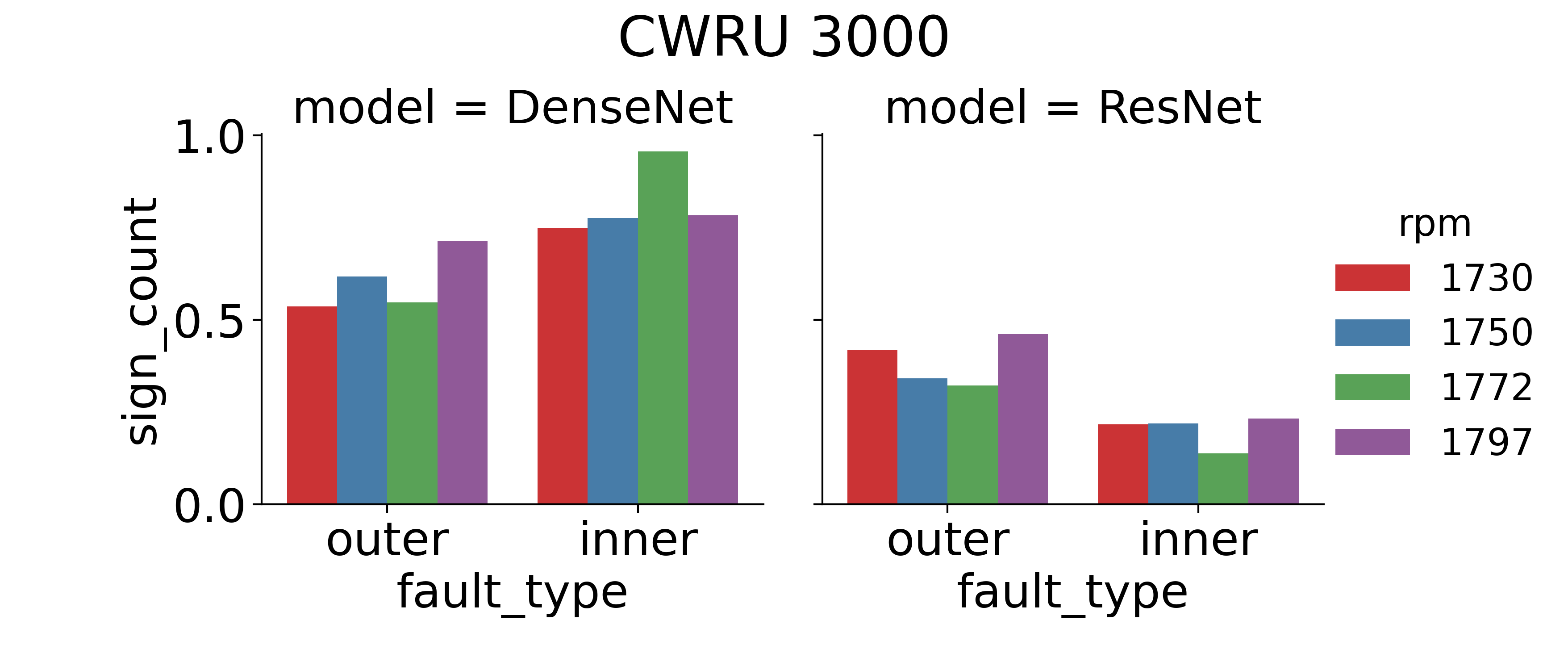}
 \includegraphics[width=0.80\columnwidth]{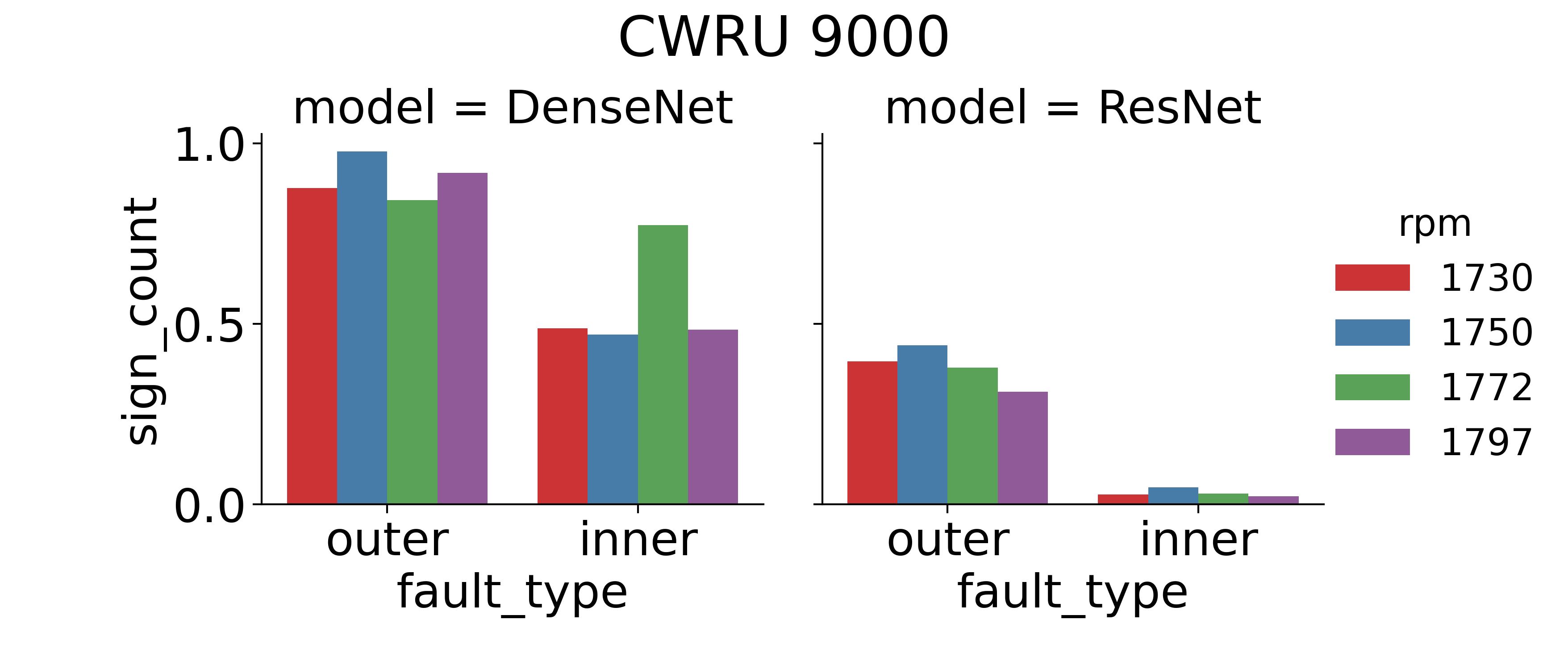}
 \includegraphics[width=0.80\columnwidth]{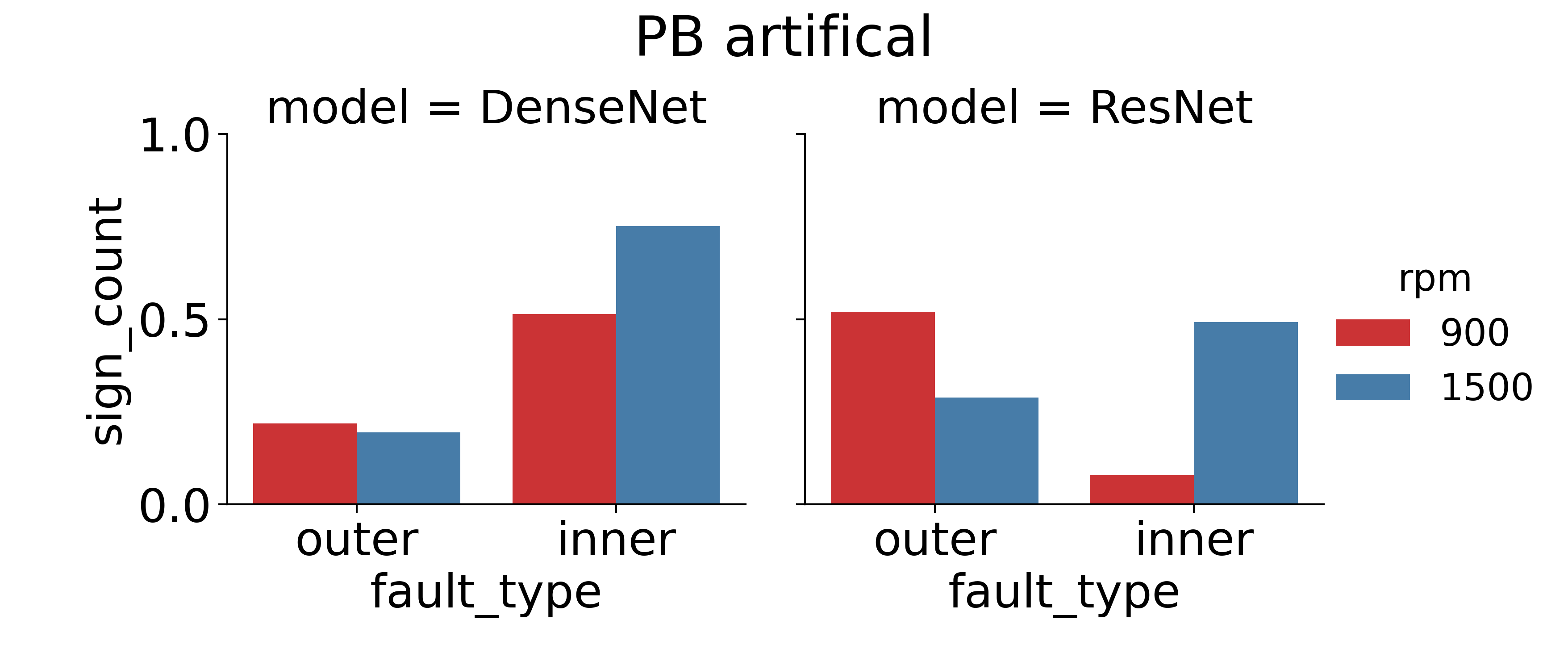}
 \includegraphics[width=0.80\columnwidth]{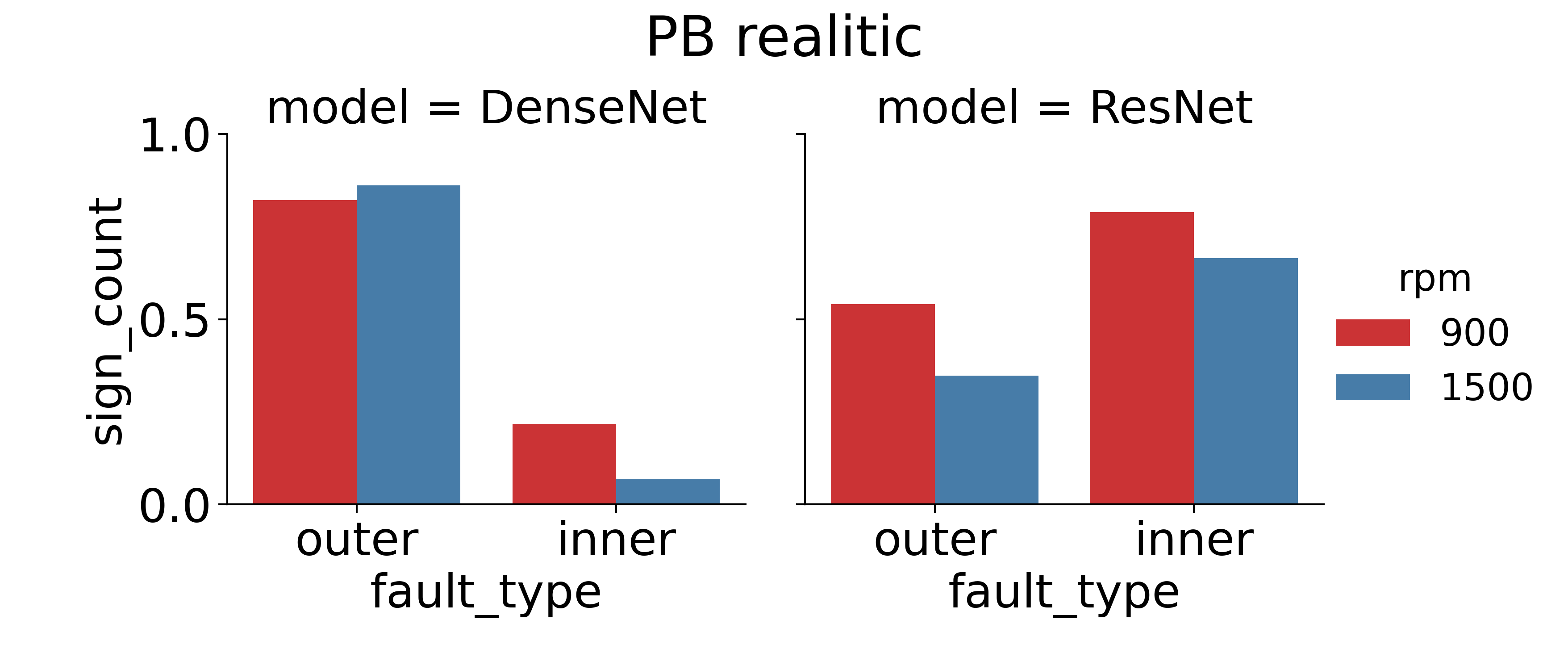}
	\caption{TCAV results of DenseNet and ResNet for all considered evaluation sets and data sources. The results imply that for predictions of models trained on CWRU data, the DenseNet architecture exhibits stronger sensitivity to vibration concepts compared to ResNet independent of the segment size. On PB artificial, the TCAV scores are inconclusive, while the significant discrepancy between scores for DenseNet on PB realistic hints at a potential model deficiency.}
	\label{fig:5}
\end{figure}
\subsection{Vibration Concept Sensitivity with TCAV}

To compute vibration concept sensitivity with TCAV we considered the following setup. For all datasets, we randomly selected $100$ signals separately for each available rotation speed (in rpm) and fault type yielding to individual evaluation datasets $\mathcal{X}_{\textit{fault type}}^{rpm}$. For each evaluation dataset, we further simulated $200$ positive and negative examples of corresponding vibration concepts as described in section IV. To establish conclusive results we repeated all experiments for each evaluation set additionally with ten different pairs of simulated example sets such that each reported result constitutes an average of ten outcomes. Remember that a crucial assumption for the applicability and faithfulness of TCAV is that the concepts of interest are linearly separable in the activation space. Otherwise, the resulting CAV is not able to faithfully express the concept and the sensitivities are unreliable. In Fig. \ref{fig:4} we plot the test performances of the classifiers trained to distinguish the concepts in the respective activation space. A consistent finding across all datasets is that the concepts are only clearly separable for DenseNet and ResNet models. This demonstrates that the applicability of TCAV for industrial prediction problems is not guaranteed and requires careful initial analysis. It also hints at the conjecture that vibration concepts are easier to distinguish for architectures that incorporate skipped connections into their design which can be of interest to model developers. However, since all models exhibit high test performance on CWRU data and DenseNet and ResNet even tend to be inferior on the PB datasets, the linear separability of the vibration concept seems not to be a necessary condition for strong model performance. To assure reliable TCAV scores, we restrict all further analyses to the two models where the separability assumption is unambiguously satisfied. The resulting TCAV scores for all evaluation settings are presented in Fig. \ref{fig:5}. On CWRU3000 and CWRU9000, the DenseNet model always attains higher scores compared to ResNet across all rotation speeds and fault types. Remember that higher sensitivity to vibration concepts is a desirable model property due to its link to existing domain knowledge. Since both models show almost identical performance on test data, the difference in TCAV scores suggests here to prefer DenseNet over ResNet for actual deployment. However, the TCAV results are inconclusive on PB datasets across model types. All evaluated predictions on PB artificial are not particularly sensitive to the relevant concepts as all sign counts are predominantly around or significantly below $50\%$. Hence, the majority of predictions on signals in the different evaluation sets do not show positive sensitivity to the presence of the relevant vibration concept. Thus, despite being able to localize concepts in their activations, the models still leverage other signal patterns to perform classification on this dataset. On PB realistic, the TCAV scores for DenseNet exhibit a strong discrepancy between fault types. Predictions on outer faults attain a score of around $80\%$ for both rotation speeds, whereas the respective scores for inner faults are substantially lower. To further investigate this issue, we depicted the corresponding confusion matrix of this architecture in Fig. \ref{fig:3} (right). Note that when presented with an outer ring vibration, the model is able to detect this fault type in $89\%$ (1013/1136) of the cases. This is in line with the high TCAV scores for outer faults, which imply that during such predictions the DenseNet is sensitive to the desirable vibration concept. However, when facing inner fault vibrations, the model is only able to correctly identify $60\%$ (772/1289). This observation matches with the low TCAV score for inner faults which suggests that the model neglects the domain knowledge that is exemplified by vibration concepts when evaluating inner faults.

\section{Conclusion}
In this work, we investigated the applicability and utility of TCAV to better understand the inner workings of different deep neural networks trained to detect bearing faults in vibration signals. For this purpose, we introduced simulated resonance vibrations with characteristic modulation as meaningful concepts that exemplify relevant physical signal properties. Our results imply that the retrieved concept-based explanation can indeed produce novel insight once the underlying assumptions are satisfied. Our analysis can be complemented by defining additional concepts of interest based on more sophisticated signal models enabling to alter and control more nuanced signal properties. We hope that our analyses can contribute to a better understanding for domain experts of how deep neural networks are solving industry-relevant prediction problems, which is crucial to increase their trustworthiness and applicability for real-world deployment. 

\bibliographystyle{IEEEtran}
\bibliography{IEEEabrv,indin.bib}

\end{document}